\documentclass[10pt,twocolumn,letterpaper]{article}

\usepackage{cvpr}
\usepackage{multirow}
\PassOptionsToPackage{table,dvipsnames}{xcolor}
\usepackage[table]{xcolor}
\usepackage{colortbl}

\definecolor{cvprblue}{rgb}{0.21,0.49,0.74}
\definecolor{rou}{HTML}{b38270}
\definecolor{mygray}{gray}{.9}
\usepackage[pagebackref,breaklinks,colorlinks,allcolors=cvprblue]{hyperref}

\def\paperID{6931} 
\def\confName{CVPR}
\def\confYear{2025}

\title{StickMotion \raisebox{-1mm}{\includegraphics[scale=0.12]{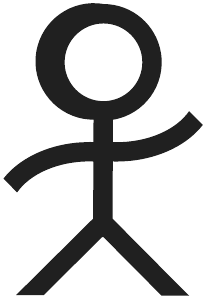}}: Generating 3D Human Motions by Drawing a Stickman}

\author{Tao Wang$^{*1}$, Zhihua Wu$^{*2}$, Qiaozhi He$^{3}$, Jiaming Chu$^{1}$, Ling Qian$^{4}$, \\ Yu Cheng$^{5}$, Junliang Xing$^{6}$, Jian Zhao$^{7, 8}$, Lei Jin$^{\dag1}$ \\
{\small $^{1}$Beijing University of Posts and Telecommunications, $^{2}$University of Science and Technology of China,} \\
{\small $^{3}$NLP Lab, School of Computer Science and Engineering, Northeastern University, Shenyang, China,} \\
{\small $^{4}$China Mobile (Suzhou) Sofware Technology Co, Ltd., $^{5}$National University of Singapore,} \\
{\small $^{6}$Tsinghua University, $^{7}$China Telecom Institute of AI, $^{8}$Northwestern Polytechnical University} \\
{\tt\small wangtao@bupt.edu.cn, wangzwhu@whu.edu.cn, qiaozhihe2022@outlook.com,} \\ 
{\tt\small  chuiaming886@bupt.edu.cn, caidunbo@cmss.chinamobile.com, e0321276@u.nus.edu,} \\
{\tt\small jlxing@tsinghua.edu.cn, jian\_zhao@nwpu.edu.cn, jinlei@bupt.edu.cn}
}

\begin{document}

\twocolumn[{%
\vspace{-5mm}
\renewcommand\twocolumn[1][]{#1}%
\maketitle
\begin{center}
    \centering
    \captionsetup{type=figure}
    \includegraphics[width=1.0\textwidth]{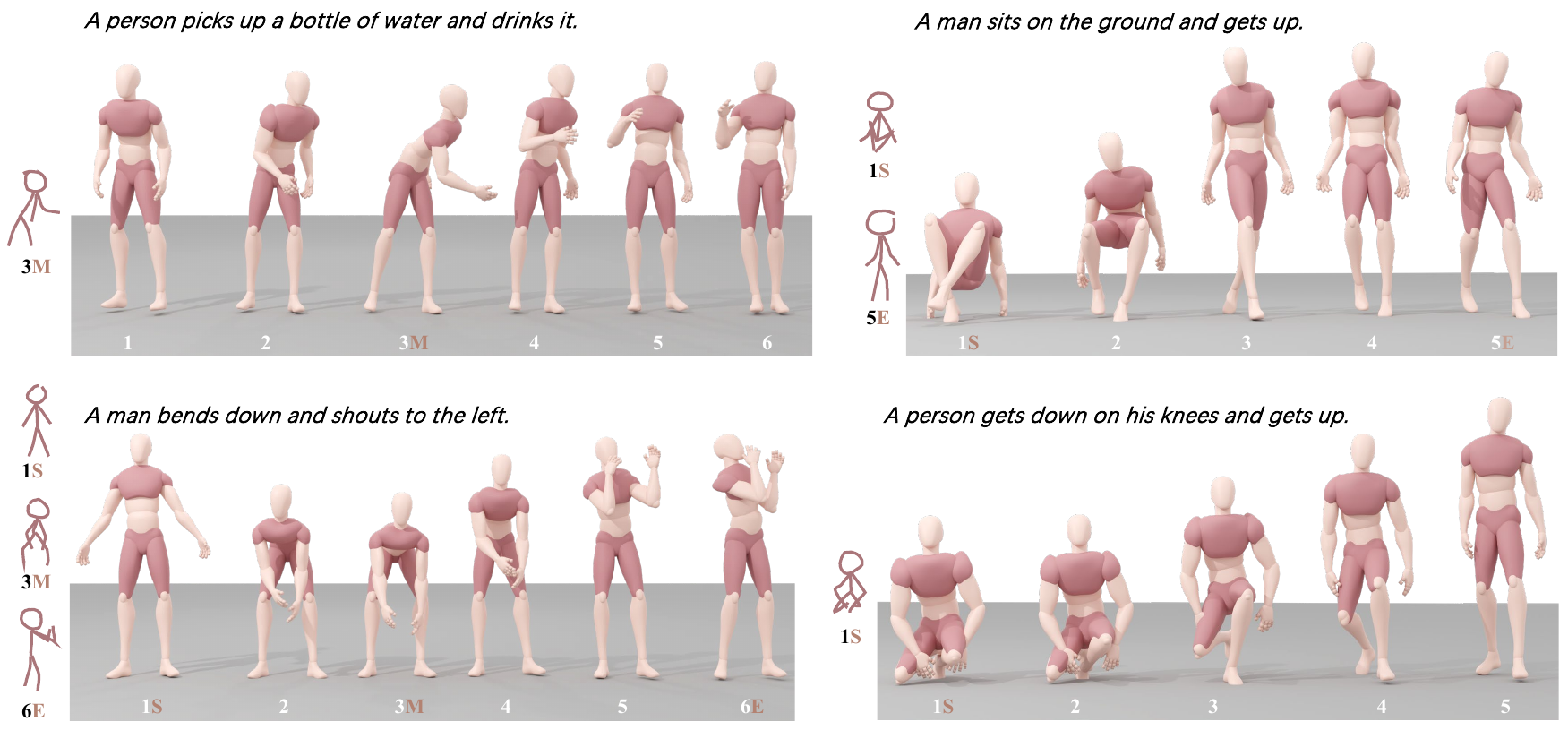}
    \captionof{figure}{Human motions generated by \textbf{StickMotion} under both stickmen and textual description conditions. The black number under the stickman denotes the index of a frame in the generated motion sequences, at which the human pose is generated with regarding the stickman. \textcolor{rou}{S}, \textcolor{rou}{M}, and \textcolor{rou}{E} denote the start, middle, and end of this motion sequence. \textbf{These above stickman figures are drawn by users.}}\label{fig:vis}
\end{center}%
}]

\renewcommand{\thefootnote}{}
\footnote{\textsuperscript{*}Both authors contributed equally. \textsuperscript{\dag}Lei Jin is corresponding author.}

\begin{abstract}
Text-to-motion generation, which translates textual descriptions into human motions, has been challenging in accurately capturing detailed user-imagined motions from simple text inputs. This paper introduces \textbf{StickMotion}, an efficient diffusion-based network designed for multi-condition scenarios, which generates desired motions based on traditional text and our proposed stickman conditions for global and local control of these motions, respectively. We address the challenges introduced by the user-friendly stickman from three perspectives:  1) \textbf{Data generation.} We develop an algorithm to generate hand-drawn stickmen automatically across different dataset formats. 2) \textbf{Multi-condition fusion.} We propose a multi-condition module that integrates into the diffusion process and obtains outputs of all possible condition combinations, reducing computational complexity and enhancing StickMotion's performance compared to conventional approaches with the self-attention module. 3) \textbf{Dynamic supervision.} We empower StickMotion to make minor adjustments to the stickman's position within the output sequences, generating more natural movements through our proposed dynamic supervision strategy. Through quantitative experiments and user studies, sketching stickmen saves users about 51.5\% of their time generating motions consistent with their imagination. Our codes, demos, and relevant data will be released to facilitate further research and validation within the scientific community.
\end{abstract}

\section{Introduction}
The task of human motion generation~\cite{zhang2023remodiffuse, zhang2024motiongpt, tevet2022motionclip} has a wide range of applications across diverse fields, including film and television production, virtual reality simulations, the gaming industry, and beyond. Specifically, the popular sub-task of motion generation, text-to-motion task, can generate natural human motion sequences based on language descriptions, freeing 3D animators from manually key-framing 3D character poses.

However,  it is evident that  a simple description such as ``A high kick forward" may not fully satisfy users' detailed imagination of the complex arm gesture shown in Fig.~\ref{fig:main}. 
Previous works~\cite{kim2023flame, zhang2024motiongpt, zhang2024finemogen, zhang2022motiondiffuse} focus on generating the fine-grained motion with complex textual descriptions. For instance, Flame~\cite{kim2023flame} allows for appending additional textual descriptions to modify the character's motion sequence based on a diffusion model. FineMoGen~\cite{zhang2024finemogen} controls the individual body parts of the 3D character through detailed descriptions. 
These works enable users to generate motion sequences that closely align with their intentions through enhanced textual descriptions and increased generation iterations. However, the user's demand for more accurate outputs necessitates more detailed textual descriptions. Based on the above, we propose a novel stickman representation to control the details of human motion sequences and mitigate the need for extensive textual descriptions.

Applying stickman condition may compromise the naturalness of the generated motion if stickman figures are directly assigned to the specified frame in the generated motion sequence, as observed in our experiments (details are presented in the supplementary materials).  Considering the ease of use, we propose that users only need to specify an approximate position for a stickman (\emph{i.e.}, start, middle, or end of the generated motion sequence). StickMotion dynamically adjusts the position of the stickman slightly based on actual conditions to maintain the naturalness of the generated motion sequence. Consequently, StickMotion should generate motion sequences by considering any combination of four inputs: textual descriptions and three stickman figures positioned at beginning, middle, and end stages of the motion sequence as presented in Fig.~\ref{fig:vis}.

Nevertheless, these above desired functionalities mainly pose three challenges for StickMotion: 
1) Data generation. Hand-drawn stickman figures are limited by the drawing style of the annotators and are time-consuming to collect. We propose the Stickman Generation Algorithm (SGA) to randomly generate stickman figures in various styles as shown in Fig.~\ref{fig:stickman}.
2) Supervision in Training. The user only needs to specify the stickman's approximate position and the network will adjust the index of the stickman near the specified position. This functionality is achieved by the proposed Dynamic Supervision as shown in Sec.~\ref{sec:loss}. 
3) Multi-Condition Fusion for the diffusion process. Previous works~\cite{chen2022re, zhang2023remodiffuse} achieve all possible combinations of two conditions via the mask operation for condition input in self-attention~\cite{vaswani2017attention, zhang2023remodiffuse} module. Still, it will introduce unnecessary computation when calculating the attention of the masked token. Moreover, the not aligned representation~\cite{sucholutsky2023getting} between stickman and text could cause the decrease of performance with the self-attention module as shown in Sec.~\ref{sec:abl_module}. Consequently, we propose an efficient Multi-Condition Module (MCM) to deal with multiple conditions as presented in Sec.~\ref{sec:fusion}.

The main contributions are summarized as follows:

\begin{enumerate}[]  
\item[\textbullet\ ] To our best knowledge, we are the first to propose a novel stickman representation for motion generation. Furthermore, we design a stickman generation algorithm to generate stickman automatically from existing datasets.
\item[\textbullet\ ] We propose the Multi-Condition Module for condition fusion in the diffusion process, which reduces computational complexity and enhances the performance of StickMotion compared to the self-attention module. 
\item[\textbullet\ ] We design a Dynamic Supervision strategy to achieve user-friendly interaction, enabling StickMotion to adjust the suitable position of the stickman during generation.
 \item[\textbullet\ ] We evaluate our proposed StickMotion on both the KIT-ML dataset and the HumanML3D dataset, demonstrating its comparable performance to state-of-the-art text-to-motion approaches. In addition, we propose a new metric named StiSim to assess the extent of bias towards the stickman figure in the generated results.
\end{enumerate}

\section{Related Work}

\noindent\textbf{Human Motion Generation.}
Human motion generation aims to generate natural sequences of human motion based on various forms of control conditions. This task can be categorized into the following types depending on the conditions. 
Motion prediction task~\cite{guo2023back, chen2023humanmac, zhang2024incorporating, ma2022progressively, wang2024gcnext} involves using previous human motion sequences as input to predict the subsequent sequences. This task can be applied to autonomous driving and social security analysis. 
Action-to-motion task~\cite{guo2020action2motion, yu2020structure, degardin2022generative, petrovich2021action, lu2022action, cervantes2022implicit} generates human motion sequences based on specified action categories, providing a more direct but coarse-grained control over human motion. 
Sound-to-motion task can be further divided into music-to-dance~\cite{gao2023dancemeld, huang2020dance, li2022danceformer, tseng2023edge} and speech-to-gesture~\cite{ao2023gesturediffuclip, ghorbani2023zeroeggs, kucherenko2019analyzing, yoon2020speech} tasks, which generate corresponding human motions or gestures in response to audio stimuli simultaneously. 
Text-to-motion task~\cite{ahuja2019language2pose, ghosh2021synthesis, tevet2022motionclip, guo2022generating, cui2024anyskill, zhang2023remodiffuse, zhang2024motiongpt, guo2024momask} generates human motion sequences from natural language descriptions like ``walk fast and turn right" or ``squat down then jump up". However, users may face challenges in precisely controlling the position of each limb with limited textual description alone. 
Additionally, there are interaction-to-motion tasks that consider interactions between humans and scenes~\cite{huang2023diffusion, hassan2021populating, lim2023mammos, liu2023revisit, xiao2023unified} / objects~\cite{diller2024cg, xu2023interdiff, gao2020interactgan, lin2023handdiffuse} / humans~\cite{cai2024digital, chopin2024bipartite, ghosh2023remos, liang2024intergen, tanaka2023role}, while incorporating generated human motions as reactions from digital lives.

\noindent\textbf{Diffusion Models.}
In recent years, significant progress has been made in applying deep learning-based generative models, particularly in diffusion models. The proposed denoising diffusion probabilistic model (DDPM)~\cite{sohl2015deep,ho2020denoising} aims to learn the process of restoring original data that has been corrupted by noise, progressively eliminating the noise during inference and resulting in final outputs that closely approximate the distribution of the original data. ADM~\cite{dhariwal2021diffusion} firstly achieves superior sample quality compared to Generative Adversarial Networks (GAN)~\cite{goodfellow2020generative} with their proposed Denoising Diffusion Implicit Model (DDIM). ADM also incorporates classifier guidance inspired by GANs to control the categories of generated content; however, this approach requires additional classifier training and is limited by its ability to ensure output quality. To address this issue, Jonathan Ho and Tim Salimans~\cite{ho2021classifier} propose a classifier-free guidance technique for improving sample quality while reducing sample diversity in diffusion models without relying on a classifier. Currently, diffusion models~\cite{cao2024survey} are employed for generating various data types such as images, videos, text, sound, time series data, \emph{etc}.

\section{Method}

\noindent\textbf{Overview.} 
StickMotion employs the stickman and the textual descriptions as input modalities. Specifically, users can provide any combination of the four support inputs, including three stickmen positioned near the start, middle, and end of their imaged motion sequence and an accompanying textual description. For instance, users can offer a textual description alongside a middle stickman or opt for a start and end stickman without additional textual context. Furthermore, StickMotion also allows controlling the bias towards generating motions based on either text or stickman conditions through the implementation of the diffusion process.

\subsection{Stickman Representation}\label{sec:stickre}
Using stickman figures obviates the necessity for extensive textual descriptions in generating desired human motion sequences, as shown in Fig.~\ref{fig:vis}. However, it is also crucial to address the challenges associated with the generation, encoding, and application of the stickman figures.

\begin{figure}[t]
\begin{center}
\includegraphics[width=0.9\linewidth]{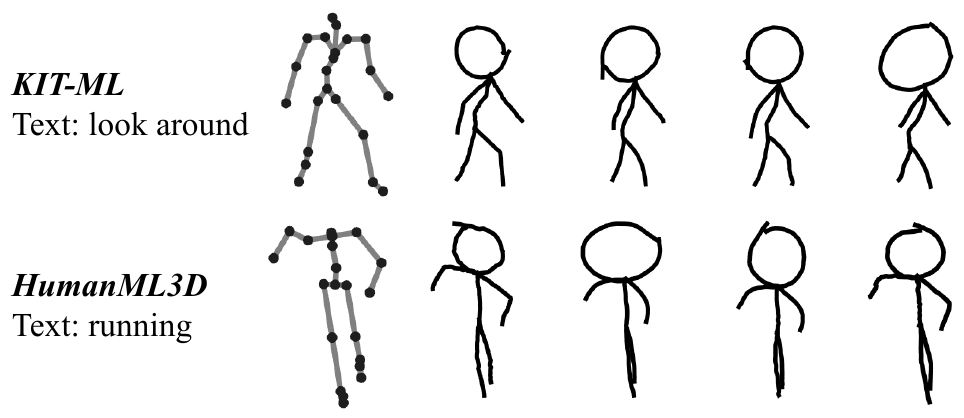}
\end{center}
\vspace{-4mm}
\caption{\small{Stickmen generated by Stickman Generation Algorithm on the KIT-ML~\cite{plappert2016kit} and HumanML3D~\cite{guo2022generating} dataset.}}
\vspace{-2mm}
\label{fig:stickman}
\end{figure}

\noindent\textbf{Stickman Generation Algorithm.}  Due to the lack of hand-draw stickman in existing datasets, we propose a Stickman Generation Algorithm (SGA) based on the 3D coordinates of human joints from existing datasets to generate the hand-draw stickman automatically. Considering the characteristics of human hand-drawing, we have taken into account the following aspects: 
   1) Stroke smoothness: The smoothness of strokes is influenced by force and individual preferences. Moreover, drawing trajectories may exhibit variations in smoothness across different devices. For instance, strokes made with a mouse tend to be more jittery than those created on an iPad.
   2) Misplacement: Inevitably, mistakes in pen placement can lead to global position deviations in these body parts.
   3) Scaling: Hand drawings focus on local details while disregarding global information, resulting in size discrepancies among different body parts.
The stickmen generated from different datasets  are shown in Fig.~\ref{fig:stickman}. Moreover, the stickman may appear similar when observing different poses from various angles, so we stipulate that the stickman should be obtained by observing the human pose from the front, \emph{i.e.}, where the line of sight is approximately perpendicular to the pelvic plane of the pose.

\noindent\textbf{Information Encoding.}
A trade-off exists between user convenience and efficient computing in handling stickman information. To accurately reconstruct the user's drawing, it is recommended to collect a minimum of 200 (estimated based on visualization) 2D coordinate points and their connectivity information when users can draw freely. However, this approach necessitates significant memory and computational resources due to the involvement of at least $200^2$ interactions among these data points through the network. To mitigate resource consumption, we propose a guideline where users should depict six one-stroke lines representing the head, torso, and four limbs in any order. Each line is individually encoded and subsequently interacted with each other using a simple transformer encoder~\cite{vaswani2017attention} to obtain an embedding for a stickman figure. This method effectively reduces computational requirements while improving the accuracy of stickman recognition.

\begin{figure*}[t]
\begin{center}
\includegraphics[width=1.0\linewidth]{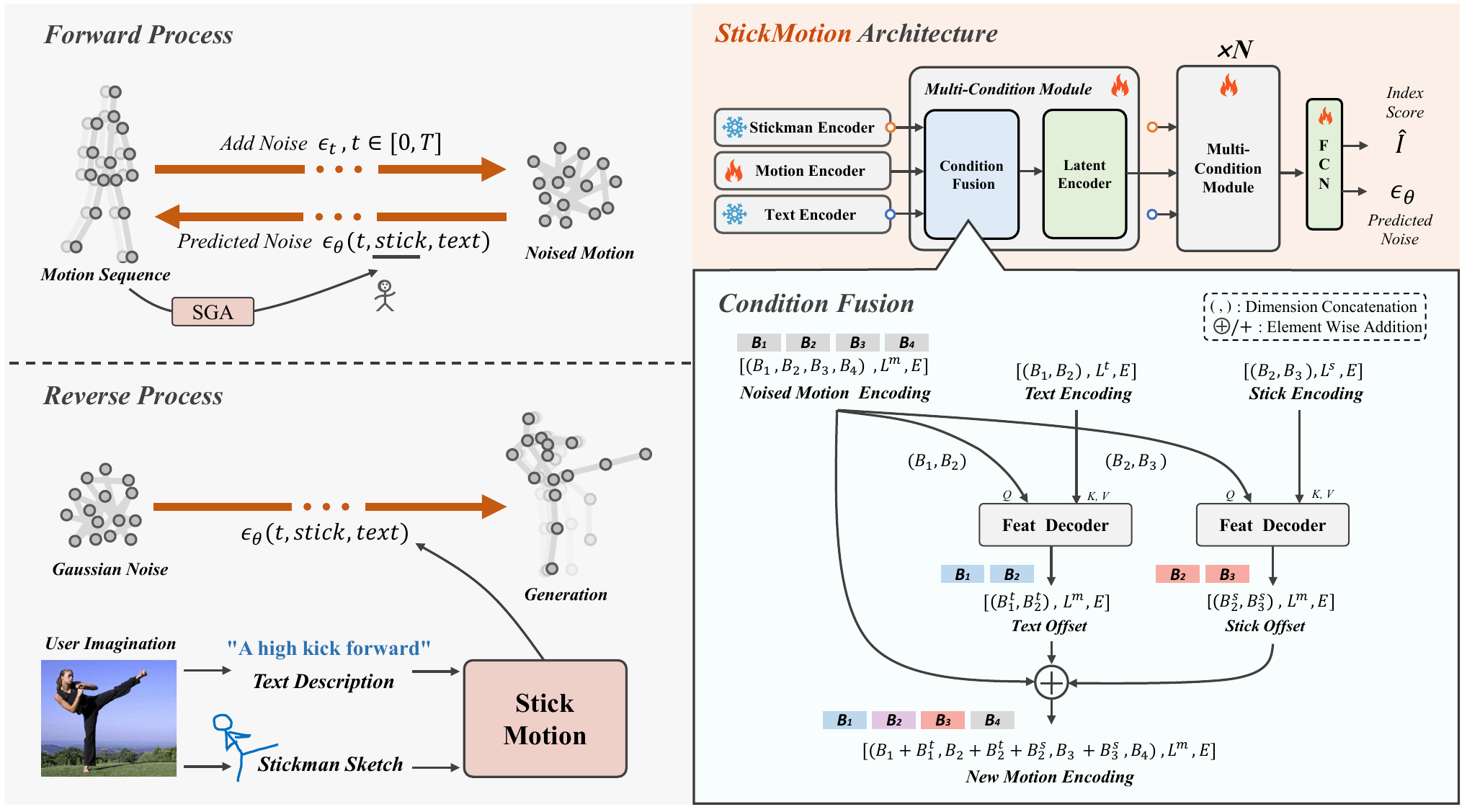}
\end{center}
\vspace{-2mm}
\caption{\small{The StickMotion framework consists of the diffusion process on the left and the network structure on the right. 
1) The diffusion process is divided into two components: the forward process and the reverse process. In the forward process, original motions are artificially augmented with Gaussian noise and fed into StickMotion to facilitate its prediction of the added noise based on text from the dataset and stickman generated by actual motion through the Stickman Generation Algorithm (SGA). In the reverse process, the user's textual descriptions and stickman figures are inputted into StickMotion, enabling the gradual generation of motion sequences with its predicted noise.
2) Regarding the structure of StickMotion, both the stickman encoder and text encoder remain frozen while other components participate in training. After encoding the input data, it undergoes multiple Multi-Condition Modules (MCM) to obtain predictions for noise, which are then utilized in generating motion sequences during the reverse process.}}
\label{fig:main}
\end{figure*}

\subsection{Diffusion-based Motion Generation}\label{sec:diffusion}
Diffusion-based works have demonstrated excellent performance in the field of human motion generation. We selected it as the base model for StickMotion since it can control the bias towards generating motions based on either textual description or stickman conditions.  
Diffusion models aim to estimate the model distribution $q_\theta(x_0)$ based on samples drawn from the motion distribution $q(x_0)$. Here, $\theta$ represents the learnable parameters in StickMotion. 

As shown in Fig.~\ref{fig:main}, the diffusion models decompose this task into a forward process and a reverse process,  which involves adding noise to actual motion and restoring the motion from noise, respectively. In the forward process, Gaussian noise $\epsilon_t \sim \mathcal{N}(0, \mathbf{I})$ is gradually  added into the actual motion $x_0$ over a time interval from $t=0$ to $t=T$, with an increasing sequence $\beta_t$ serving as its weight, as shown in Equ.~\ref{eq:forward}. 
\begin{equation} 
\begin{aligned}
&q(\mathbf{x}_{1:T}|\mathbf{x}_0) = \prod_{t=1}^Tq(\mathbf{x}_t|\mathbf{x}_{t-1}),\\
&q(\mathbf{x}_t|\mathbf{x}_{t-1}) = \mathcal{N}(\mathbf{x}_t;\sqrt{\alpha_t}\mathbf{x}_{t-1}, (1-\alpha_t)\mathbf{I}).
\end{aligned}\label{eq:forward}
\end{equation}
This formula is equivalent to $\mathbf{x}=\sqrt{\bar{\alpha}_{t}}\mathbf{x}_{0}+\sqrt{1-\bar{\alpha}_{t}}\epsilon_t$, where $\bar{\alpha}_t=\prod_{s=1}^t\alpha_s$. In this way, we can directly  get $x_t$ from $x_0$ without the intermediate variables. The forward process above is used for the training of StickMotion. To achieve a balance between stickman and text conditions for the final generation during inference, we employ the classifier-free diffusion guidance~\cite{ho2022classifier} for the diffusion process, and the supervision formula is set as 
\begin{equation}
\begin{aligned}
\mathbb{E}_{\epsilon_t,t,x_0}[||\epsilon_t-\epsilon_\theta(\mathbf{x}_t,t,L,C(stick), C(text))||^2],
\end{aligned}\label{eq:motion_loss}
\end{equation}
where $L$ represents the length of motion sequences. The stickman condition considered in training is denoted as $C(stick)$ with a probability of $p^c_{stick}$, while the text condition is represented as $C(text)$. $\epsilon_\theta(\cdot)$ denotes the predicted noise from StickMotion. We set $p^c_{stick}=p^c_{text}=0.7$ during the training process.

In the reverse process, the well-trained StickMotion gradually removes the noise from $X_T$ to get the real motion sequence. According to DDPM~\cite{ho2020denoising}, $x_{t-1}$ can be sampled from $p_\theta(x_{t-1}|x_t)$ by the formula 
\begin{equation} 
\begin{aligned}
x_{t-1}=\frac{1}{\sqrt{\alpha_{t}}}(x_{t}-\frac{1-\alpha_t}{\sqrt{1-\bar{\alpha}_{t}}}\epsilon_{\theta})+\sigma_{t},
\end{aligned}\label{eq:reverse}
\end{equation}
where $\epsilon_{\theta}$ is the predicted noise from StickMotion, and $\sigma_{t}$ is a weighted Gaussian noise. In the reverse process from T to 0, we can get outputs that are biased to different combinations of conditions as follows. 
\begin{equation}
\begin{aligned}
\hat{\epsilon_{\theta}} =\ &w_1\cdot\epsilon_{\theta}(stick, text)+w_2 \cdot \epsilon_{\theta}(stick,)+\\
&w_3 \cdot \epsilon_{\theta}(text,)+w_4\cdot\epsilon_{\theta}(,).
\end{aligned}\label{eq:mixture}
\end{equation}

Adhering to the principle $w_1+w_2+w_3+w_4=1$~\cite{ho2021classifier} that preserves output statistics, we propose an efficient condition mixture by considering the characteristics of stickman and text conditions. Initially, the approximate motion sequence is determined during the beginning stage of time interval $t \in [T, T/10]$, the condition mixture follows the formula $(w_1=w$, $w_2=\hat{w}$, $w_3=w-\hat{w}, w_4=1-2\cdot w)$. Here, 1) constant $w>1$~\cite{ho2021classifier} adjusts the condition sampling strength; 2) $w_1=w$ ensures that the fusion of stickman and text is harmonious. 3) $p(\hat{w}=w)+p(\hat{w}=0)=1$, with $p(\hat{w}= w)$ controlling the preference of generated motion for the stickman condition. 4) $w_4=1-2\cdot w$ controls the constant distribution of the output. In the final stage ($t \in [T/10, 0]$), we set $(w_1=1, w_{2,3,4}=0)$ to use all conditions to further refine the preliminary result from the beginning stage. Finally, a motion sequence corresponding to the stickman condition and text condition is generated in the reverse process. More details can be found in Sec.~\ref{sec:abl_mixture} and the supplementary materials.

\subsection{Architecture of StickMotion}\label{sec:network}
The diffusion model offers StickMotion a straightforward training and controllable generation process. However, designing a network architecture that efficiently handles both stickman and text conditions for the diffusion process is equally crucial. As depicted in the right section of Fig.~\ref{fig:main}, StickMotion comprises three input encoders and Multi-Condition Modules (MCM) to generate the motion sequence's final predicted noise and stickman index score. The input encoders transform the noisy motion, text, and stickman into vectors, which are subsequently fed into  MCMs to produce the final outputs under multiple condition combinations, \emph{i.e.},  {\small$(text), (text, stick), (stick),$} and {\small$()$}.

\noindent\textbf{Input.}
The input data consists of noisy motion sequences, stickman figures, and text, which are encoded to the shape of $[L^m,E], [L^s,E]$ and $[L^t,E]$, respectively. Here, $L$ denotes the length of the input encoding, and $E$ represents the dimension of each token in the encoding. We employ a simple linear layer to encode the motion sequences. We utilize CLIP ViT-B/32~\cite{clip}, which consists of 154M parameters for text encoding. Regarding the stickman figures, we leverage a standard transformer encoder with six encoder layers. However, based on experimental findings, it has been observed that pre-training and freezing the stickman encoder significantly enhances StickMotion's performance. Consequently, we train a pair of auto-encoder models for the stickman condition: a stickman encoder and a feature-to-pose decoder. The stickman encoder encodes the stickman into a feature embedding. At the same time, the feature-to-pose decoder reconstructs the original pose from this feature vector to ensure the preservation of pose information.

\noindent\textbf{Multi-Condition Module.}\label{sec:fusion}
Condition combinations are essential for the diffusion process to fuse these conditions. These combinations are implemented through the Multi-Condition Module (MCM) in StickMotion.  Within each MCM, a Condition Fusion module is utilized to incorporate stickman and text conditions into the motion encoding in the latent space. Subsequently, the modified motion encoding undergoes re-encoding by the Latent Encoder for further fusion.
Specifically, we partition all data along the batch dimension into four segments ({\small$B_1, B_2, B_3, B_4$}), representing four combinations of text and stickman conditions,  \emph{i.e.},  {\small$(text), (text, stick), (stick),$} and {\small$()$}. 
As shown in Fig.~\ref{fig:main}, in the Condition Diffusion, two standard transformer decoder layers named Feat Decoder consider solely the text input and stickman input for batches ({\small$B_1,B_2$}) and ({\small$B_2,B_3$}), respectively.
By summing up these predicted offsets with their corresponding motion encoding along the batch dimension, we obtain new motion encoding for three condition combinations with only two Feat Decoders as shown in Fig.~\ref{fig:main}. Sequently, the new motion encoding undergoes re-encoding in the Latent Encoder to facilitate further information fusion.  
Compared to conventional approaches with self-attention module that accomplish all condition combinations by the partial mask operation for condition inputs, MCM reduces computational complexity and enhances StickMotion's performance as demonstrated in Sec.~\ref{sec:abl_module}. Additionally, we employ efficient attention~\cite{shen2021efficient, zhang2023remodiffuse} for both feat decoder and latent encoder to further reduce computational complexity.

\noindent\textbf{Output.}
StickMotion outputs predicted noise for noisy motions and an index score for the position of input stickmen. The index scores are crucial for user interaction and training supervision of stickmen: 1) The index score can explicitly supervise and minimize the distance between the dynamically assigned pose and stickman during training, thereby making the results closer to the stickman as presented in Sec.~\ref{sec:loss}. 2) The index score indicates the position of the stickman in the generated motion sequences, enabling users to decide whether to adjust the generation towards the stickman figures or textual descriptions freely.

\subsection{Dynamic Supervision}\label{sec:loss}
Different from previous works, StickMotion faces an additional challenge in determining the stickman indexes of the generated motion sequence. Allowing users to specify the index of each stickman manually would introduce an extra burden on them to consider suitable indexes, potentially resulting in less naturally generated results due to improper positioning of multiple stickman figures. To address this issue, we have designated three positions, start, middle, and end, from which users can choose. The network will dynamically adjust the indexes of stickmen near these specified positions to optimize  the generation's naturalness and adherence to the textual description. 

Specifically,  we first randomly sample human poses at the beginning, middle, and end positions within the ranges {\small$[0, 1/4\cdot L]$}, {\small$[3/8\cdot L, 5/8\cdot L]$}, and {\small$[3/4\cdot L, L]$} ({\small$L$} represents the length of motion sequence), respectively. Then, we randomly create a mask {\small$M$} for each stickman to mimic users' random input.  The overall loss function is divided into two parts as follows.
\begin{equation} 
\begin{aligned}
 \mathcal{L}_{\rm index} &= M\cdot\sum_{l=0}^{L} softmax(\hat{I}_l) \cdot ||\hat{x}_l(stick,*) - x_i||^2, \\
 \mathcal{L}_{\rm motion} &= \sum_{l=0}^{L} ||\hat{x}_l(*,*) - x_l||^2, \\
 \mathcal{L}_{\rm total} &= \mathcal{L}_{\rm index}^{start} + \mathcal{L}_{\rm index}^{middle} + \mathcal{L}_{\rm index}^{end} + \mathcal{L}_{\rm motion},
\end{aligned}\label{eq:samp}
\end{equation}

The predicted index score for the stickman at frame $l$ in the generated motion sequence is denoted as $\hat{I}_l$. The highest score at frame $l$ indicates that StickMotion considers the input stickman to belong to that specific frame. $\hat{x}(stick, *)$ refers to the generated motion sequence under condition combinations of $(stick, text)$ and $(stick,)$. Predicted poses at frame $l$ with all condition combinations are denoted as $\hat{x}_t(*,*)$. Here, $x_l$ represents the pose of the real motion sequence at frame $l$, while a stickman is generated by pose $x_i$ from the ground-truth motion sequence through the Stickman Generation Algorithm. In this way, $\mathcal{L}_{\rm index}$ ensures that the assigned pose $\hat{x}_l(stick,*)$, which has the highest score in the output sequence, is supervised by ground-truth pose $x_i$. The supervisions for start, middle, and end stickmen are denoted as $\mathcal{L}_{\rm index}^{start}$, $\mathcal{L}_{\rm index}^{middle}$, and $\mathcal{L}_{\rm index}^{end}$ respectively. The supervision for diffusion process in StickMotion is represented by $\mathcal{L}_{\rm motion}$ as shown in Equ.~\ref{eq:motion_loss}.

\begin{table*}[t]
\renewcommand\arraystretch{1.3}
\setlength{\tabcolsep}{1.6mm}
\caption{\small{Comparison on the HumanML3D test set. We mark the best result as \colorbox{red!15}{red} and the second best one as \colorbox{blue!15}{blue}.}}
\vspace{-2mm}
\centering
\begin{tabular}{lccccccc}
\hline 
\multirow{2}{*}{Methods} & \multicolumn{3}{c}{R Precision $\uparrow$} & \multirow{2}{*}{FID $\downarrow$} & \multirow{2}{*}{MM Dist $\downarrow$} & \multirow{2}{*}{Diversity $\uparrow$} & \multirow{2}{*}{MultiModality $\uparrow$} \\ 
& Top1 & Top2 & Top3 & & & & \\
\hline Real motions &  0.511$^{ \pm .003}$  &  0.703$^{ \pm .003}$  &  0.797$^{ \pm .002}$  &  0.002$^{ \pm .000}$  &  2.974$^{ \pm .008}$  &  9.503$^{ \pm .065}$  & - \\
\hline 
Guo et al.~\cite{guo2022generating} &  0.457$^{ \pm .002}$  &  0.639$^{ \pm .003}$  &  0.740$^{ \pm .003}$  &  1.067$^{ \pm .002}$  &  3.340$^{ \pm .008}$  &  9.188$^{ \pm .002}$  &  2.090$^{ \pm .083}$  \\
MDM~\cite{tevet2022humanmotiondiffusionmodel} & - & - &  0.611$^{ \pm .007}$  &  0.544$^{ \pm .044}$  &  5.566$^{ \pm .027}$  &\cellcolor{blue!15}   9.559$^{ \pm .086}$  &\cellcolor{red!15}   2.799$^{ \pm .072}$  \\
MotionDiffuse~\cite{zhang2022motiondiffuse} &  0.491$^{ \pm .001}$  &  0.681$^{ \pm .001}$  &  0.782$^{ \pm .001}$  &  0.630$^{ \pm .001}$  &  3.113$^{ \pm .001}$  &  9.410$^{ \pm .049}$  &  1.553$^{ \pm .042}$  \\
T2M-GPT~\cite{zhang2023generating} &  0.491$^{ \pm .003}$  &  0.680$^{ \pm .003}$  &  0.775$^{ \pm .002}$  &  0.116$^{ \pm .004}$  &  3.118$^{ \pm .011}$  & \cellcolor{red!15}  9.761$^{ \pm .081}$  &  1.856$^{ \pm .011}$  \\
ReMoDiffuse~\cite{zhang2023remodiffuse} & \cellcolor{blue!15}  0.510$^{ \pm .005}$  &\cellcolor{blue!15}   0.698$^{ \pm .006}$  &\cellcolor{blue!15}   0.795$^{ \pm .004}$  &\cellcolor{red!15}   0.103$^{ \pm .004}$  &\cellcolor{blue!15}   2.974$^{ \pm .016}$  &  9.018$^{ \pm .075}$  &  1.795$^{ \pm .043}$  \\\hline
StickMotion (Ours) &\cellcolor{red!15}   0.518$^{ \pm .007}$  &\cellcolor{red!15}   0.702$^{ \pm .003}$  &\cellcolor{red!15}   0.797$^{ \pm .005}$  &\cellcolor{blue!15}   0.107$^{ \pm .003}$  &\cellcolor{red!15}   2.953$^{ \pm .021}$  &  9.239$^{ \pm .066}$  &  \cellcolor{blue!15} 2.256$^{ \pm .051}$ \\
\hline
\end{tabular}\label{tab:t2m}
\end{table*}

\begin{table*}[t]
\renewcommand\arraystretch{1.3}
\setlength{\tabcolsep}{1.6mm}
\caption{\small{Comparison on the KIT-ML test set. }}
\vspace{-2mm}
\centering
\begin{tabular}{lccccccc}
\hline 
\multirow{2}{*}{Methods} & \multicolumn{3}{c}{R Precision $\uparrow$} & \multirow{2}{*}{FID $\downarrow$} & \multirow{2}{*}{MM Dist $\downarrow$} & \multirow{2}{*}{Diversity $\uparrow$} & \multirow{2}{*}{MultiModality $\uparrow$} \\ 
& Top1 & Top2 & Top3 & & & & \\
\hline Real motions &  0.424$^{ \pm .005}$  &  0.649$^{ \pm .006}$  &  0.779$^{ \pm .006}$  &  0.031$^{ \pm .004}$  &  2.788$^{ \pm .012}$  &  11.08$^{ \pm .097}$  & -\\
\hline 
Guo et al.~\cite{guo2022generating} &  0.370$^{ \pm .005}$  &  0.569$^{ \pm .007}$  &  0.693$^{ \pm .007}$  &  2.770$^{ \pm .109}$  &  3.401$^{ \pm .008}$  &  10.91$^{ \pm .119}$  &  1.482$^{ \pm .065}$  \\
MDM~\cite{tevet2022humanmotiondiffusionmodel} & - & - &  0.396$^{ \pm .004}$  &  0.497$^{ \pm .021}$  &  9.191$^{ \pm .022}$  &  10.847$^{ \pm .109}$  & \cellcolor{red!15} 1.907$^{ \pm .214}$  \\
MotionDiffuse~\cite{zhang2022motiondiffuse} &  0.417$^{ \pm .004}$  &  0.621$^{ \pm .004}$  &  0.739$^{ \pm .004}$  &  1.954$^{ \pm .062}$  &  2.958$^{ \pm .005}$  & \cellcolor{red!15} 11.10$^{ \pm .143}$  &  0.730$^{ \pm .013}$  \\
T2M-GPT~\cite{zhang2023generating} &  0.416$^{ \pm .006}$  &  0.627$^{ \pm .006}$  &  0.745$^{ \pm .006}$  &  0.514$^{ \pm .029}$  &  3.007$^{ \pm .023}$  &  10.921$^{ \pm .108}$  & \cellcolor{blue!15} 1.570$^{ \pm .039}$  \\
 ReMoDiffuse~\cite{zhang2023remodiffuse} & \cellcolor{blue!15} 0.427$^{ \pm .014}$  &  
\cellcolor{blue!15} 0.641$^{ \pm .004}$  & \cellcolor{blue!15} 0.765$^{ \pm .055}$  & \cellcolor{blue!15} 0.155$^{ \pm .006}$  & \cellcolor{blue!15} 2.814$^{ \pm .012}$  &  
 10.80$^{ \pm .105}$  &  1.239$^{ \pm .028}$  \\ \hline
 StickMotion (Ours) &\cellcolor{red!15} 0.430$^{ \pm .017}$  & \cellcolor{red!15} 
 0.654$^{ \pm .010}$  & \cellcolor{red!15} 0.775$^{ \pm .043}$  &  \cellcolor{red!15} 0.141$^{ \pm .008}$  & \cellcolor{red!15} 2.763$^{ \pm .018}$  & \cellcolor{blue!15} 10.94$^{ \pm .178}$  &  1.457$^{ \pm .033}$ \\
 
\hline
\end{tabular}\label{tab:kit}
\end{table*}

\section{Experiments}
\subsection{Experiment Settings}

\noindent\textbf{Dataset and Metrics.}\label{metric}
We conduct experiments on two prominent datasets of human motion generation, namely the KIT-ML dataset~\cite{plappert2016kit} and the HumanML3D dataset~\cite{guo2022generating}. The same evaluation of Guo et al.~\cite{guo2022generating} is adopted, so we can comprehensively compare with existing text-to-motion methods. This evaluation  involves encoding the input text and generated motion sequence into embeddings through pre-trained contrastive quantitative assessment models, and then participate in the calculating the following metrics. 
\textit{R Precision.} Given a prediction motion sequence, its text and 31 other irrelevant texts from the test set are combined into a set, and the Top-k accuracy between the motion and text set is calculated after passing through the pre-trained motion-text contrastive models. 
\textit{Frechet Inception Distance (FID).} Motion embeddings are generated by ground-truth and generated motion sequences through contrastive models. Then, the difference between the two batches of embedding distribution is calculated. FID is related to generation quality but is limited by the performance of contrastive models.
\textit{Multimodal Distance (MM Dist).} The Euclidean distance between the motion embedding and its text embedding.
\textit{Diversity.} The variability and richness of the generated motion sequences.
\textit{Multimodality.} The variance of generated motion sequences given a specified text.

\noindent\textbf{Implementation Details.}
StickMotion is trained with 4 A800 GPUs with a batch size of 1024, while 80 dataloader workers are adopted to generate stickmen through SGA. For the diffusion process, we set the noise steps $T=1000$. And $\alpha_t, t\in [0,T]$ ranges from 0.9999 to 0.9800. The trainable StickMotion model comprises four MCMs for the KIT-ML dataset and five MCMs for the HumanML3D dataset, with parameter counts of 43M and 62M respectively.

\subsection{Quantitative Analysis}

\noindent\textbf{Comparison with SOTA Methods.} We follow the same evaluation of  text-to-motion methods to demonstrate the performance of StickMotion based on both the KIT-ML dataset and the HumanML3D dataset. Additionally, the generation approach of the stickman used in the evaluation is similar to that of the training process as described in Sec.~\ref{sec:loss}, and all stickmen in three positions of the ground-truth sequences are utilized. In contrast to conventional approaches, StickMotion  requires alignment with both textual descriptions and stickman figures. As mentioned in Sec.~\ref{sec:abl_mixture}, these two conditions have an adversarial relationship that can negatively impact StickMotion's performance. Hence, we adjusted the reverse process mentioned in Sec.~\ref{sec:diffusion} by setting $p(\hat{w}= w)=20\%$ and $(w_1=1,w_2=0,w_3=0,w_4=0)$ to bias the generated results towards the stickman condition. Our approach shows excellent performance compared to previous text-to-motion works as shown in Tab.~\ref{tab:t2m} and Tab.~\ref{tab:kit}. However, it is crucial to assess how much stickman information was incorporated into the generated motions. Therefore, we propose a new metric called Stickman Similarity, which will be discussed as follows.

\noindent\textbf{New Metric: Stickman Similarity (StiSim)}. 
In the evaluation, the stickman is generated by sampling a pose from the ground-truth sequence, while the corresponding poses in the generated sequences are determined by the index scores generated from StickMotion. Thus, the Euclidean distance (referred to as \textit{StickmanDistance}) between the sampled ground-truth pose and the generated pose with highest index score indicates how much StickMotion has learned from the input stickman. Moreover, we use \textit{MeanDistance} to denote the mean of Euclidean distances between sampled ground-truth poses and all generated poses. Therefore, \textit{StickmanSimilarity} (StiSim) can be calculated as 1 - (\textit{StickmanDistance} / \textit{MeanDistance}). For HumanML3D dataset~\cite{guo2022generating} and KIT-ML dataset~\cite{plappert2016kit}, values for \textit{StickmanDistance}, \textit{MeanDistance}, and \textit{StickmanSimilarity} are (0.24, 0.41, 41.5\%) and (0.27, 0.47, 42.6\%), respectively.

\subsection{Ablation Study}

\noindent\textbf{Analysis on StickMotion Modules.}\label{sec:abl_module}
We first conduct ablation experiments on the Condition Fusion and Latent Encoder modules in the Multi-Condition Model (MCM) using the KIT-ML dataset. We replace Condition Fusion with conventional self-attention~\cite{vaswani2017attention} while ensuring that MCM still supports multi-condition combinations. Additionally, we calculate the Flops of a MCM with a batch size ({\small$B_1=B_2=B_3=B_4=256$}). As shown in Tab.~\ref{tab:abl_train}, combining Condition Fusion and Latent Encoder achieves better performance compared to conventional approaches with self-attention module for condition combinations through mask operations, which introduce unnecessary computations when calculating attention for masked conditions.

\begin{table}[t]
\centering
\caption{\small{Ablation study on Condition Fusion and Latent Encoder for the training / forward process.}}
\small
\renewcommand\arraystretch{1.2}
\setlength{\tabcolsep}{0.8mm}
\begin{tabular}{cccccc}
\toprule
 {\small Self-Attention} & {\small ConditionFusion} & {\small LatentEncoder} & FID$\downarrow$ & TFlops$\downarrow$    \\ \toprule
\rowcolor{mygray}  $\surd$  &     &     &  0.31 &0.46  \\
 $\surd$   &     &    $\surd$   & 0.18   &0.71 \\
\rowcolor{mygray}    &  $\surd$   &     & 0.25   &\textbf{0.28} \\
  &  $\surd$   &    $\surd$   & \textbf{0.14}   &0.43 \\
\bottomrule
\end{tabular}\label{tab:abl_train}
\vspace{-3mm}
\end{table}

\noindent\textbf{Analysis on Condition Mixture.}\label{sec:abl_mixture}
As shown in Equ.~\ref{eq:mixture}, the condition mixture controls how to mix the outputs based on four combinations of stickman and text conditions, resulting in a bias towards either stickman or text in the generated output. (The combinations of three stickmen in different positions will be discussed next, since we only focus on the these conditions at a higher level in this section.) As discussed in Sec.~\ref{sec:diffusion}, $p(\hat{w}= w)$ determines the extent to which the coarse generations are biased towards the stickman condition during the initial stage of the reverse process. Here $(w_1,w_2,w_3,w_4)$ refines these coarse generations based on these condition combinations in the final stage. According to Tab.~\ref{tab:mixture}, there exists an adversarial relationship between stickman and text conditions.

\begin{table}[t]
\centering
\caption{\small{Ablation study on the condition mixture for the inference / reverse process.}}
\small
\renewcommand\arraystretch{1.2}
\setlength{\tabcolsep}{3.5mm}
\begin{tabular}{cccccc}
\toprule
 $p(\hat{w}= w)$ & $(w_1,w_2,w_3,w_4)$ & FID$\downarrow$ & StiSim$\uparrow$    \\ \toprule
\rowcolor{mygray} 50\%   &   (0, 1, 0, 0)  &  0.147   & 44.8\%   \\
 50\%   &   (0, 0, 1, 0)   &  \textbf{0.135}   & 26.1\%  \\
\rowcolor{mygray} 80\%    & (1, 0, 0, 0)  & 0.146     &    \textbf{51.3\%} \\
 20\%  &  (1, 0, 0, 0)  &  0.141 &  42.6\% \\
\bottomrule
\end{tabular}\label{tab:mixture}
\vspace{-3mm}
\end{table}

\begin{figure}[t]
\begin{center}
\includegraphics[width=1.0\linewidth]{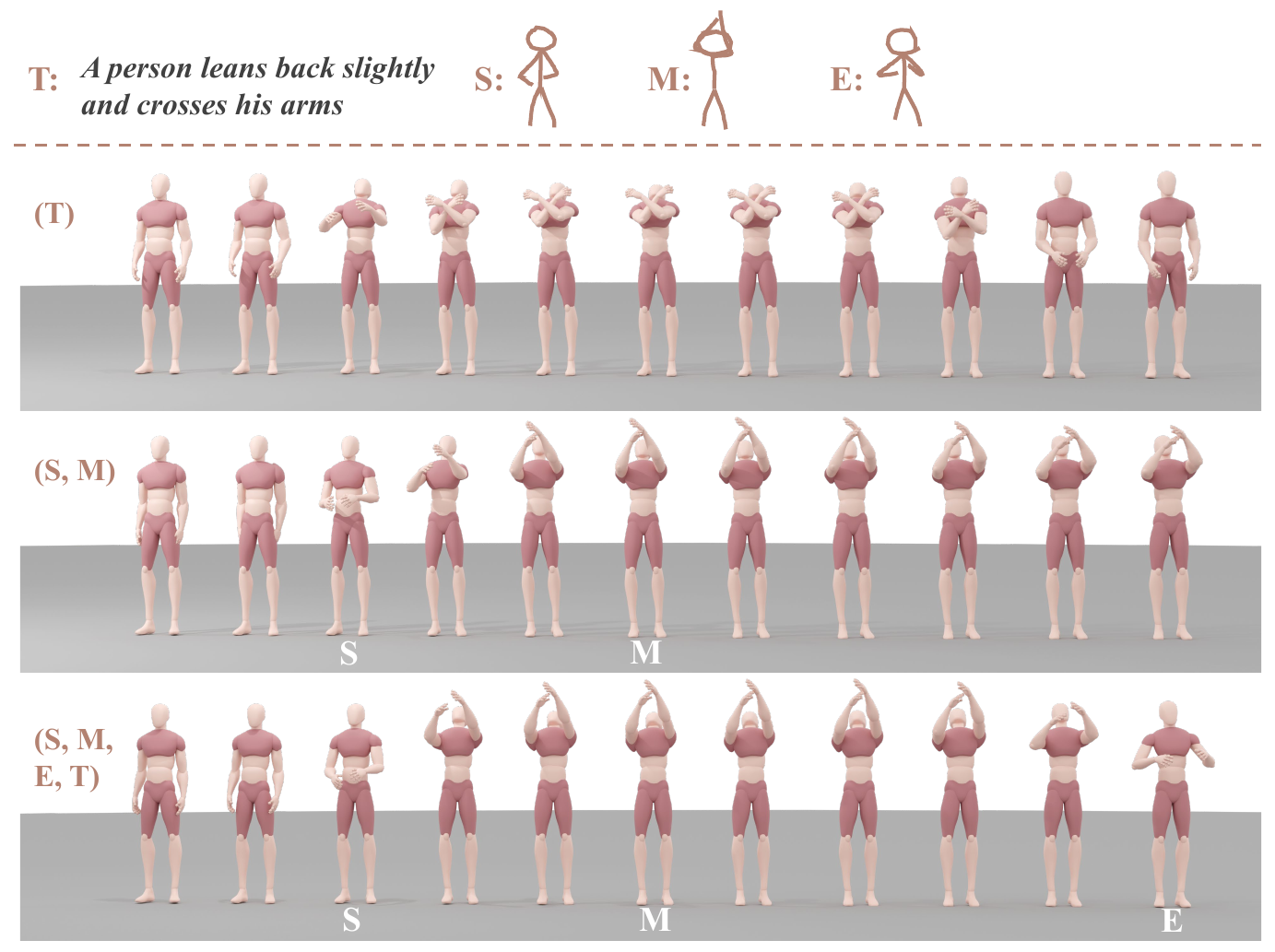}
\end{center}
\vspace{-4mm}
\caption{\small{Visualization with various input combinations.}}
\vspace{-2mm}
\label{fig:ablation}
\end{figure}

\noindent\textbf{Analysis on Input Combination.}
Users can provide diverse combinations of inputs, encompassing the textual description and three stickmen positioned at different stages (start, middle, or end) of the motion sequence. The generation quality of those combinations under the settings of $p(\hat{w}= w)=20\%$ and $(w_1=1,w_2=0,w_3=0,w_4=0)$ can be observed in Tab.~\ref{tab:abl_test}. It should be noticed that the text description has the most significant impact on the performance of StickMotion because the text-to-motion evaluation solely takes into account the semantic information of the generated motion sequences. We also show a visualization of partial combinations in Fig.~\ref{fig:ablation}.

\definecolor{mygray}{gray}{.9}
\begin{table}[t]
\centering
\caption{\small{Ablation study on combinations of conditions in the generate process. There are four inputs, including stickmen in the position of ``Start'', ``Middle'', and ``End'' along with the textual description.  }}
\small
\renewcommand\arraystretch{1.2}
\setlength{\tabcolsep}{4.9mm}
\begin{tabular}{cccccc}
\toprule
 Start & Middle & End & Text & FID$\downarrow$    \\ \toprule
\rowcolor{mygray}    &     &     &   &4.292  \\
 $\surd$ &     &    &    &3.291\\
\rowcolor{mygray}    &   $\surd$  &     &   &3.132  \\
  &     &   $\surd$ &    &3.450\\
\rowcolor{mygray}    &     &     & $\surd$  &0.164  \\
 $\surd$ &     &   $\surd$ &    &3.548\\
\rowcolor{mygray}  $\surd$  &  &   $\surd$  & $\surd$   &0.189  \\
 &   $\surd$  &  &  $\surd$  &0.157\\
\rowcolor{mygray}  $\surd$  &    $\surd$ &   $\surd$  &   &2.782  \\
 $\surd$ &   $\surd$  &  $\surd$  &  $\surd$  &\textbf{0.141}\\
\bottomrule
\end{tabular}\label{tab:abl_test}
\vspace{-3mm}
\end{table}

\begin{figure}[t]
\begin{center}
\includegraphics[width=1.0\linewidth]{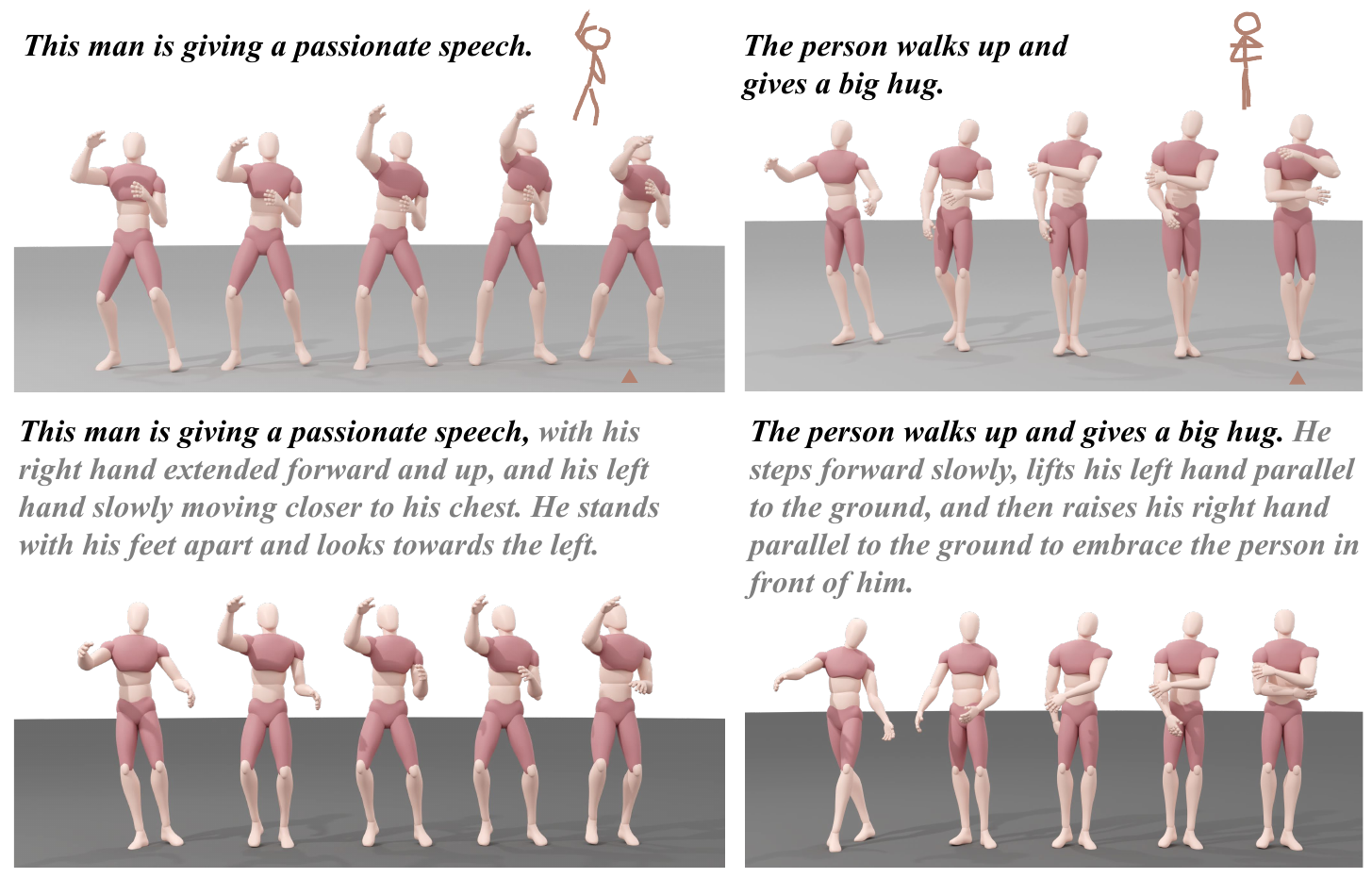}
\end{center}
\vspace{-4mm}
\caption{\small{Comparison between overall description \& stickman for StickMotion (above) and detailed  description for ReMoDiffuse (below).}}
\vspace{-2mm}
\label{fig:user}
\end{figure}

\noindent\textbf{User Study.} We recruited 20 unrelated volunteers to participate in the user study of StickMotion and the traditional text-to-motion work ReMoDiffuse~\cite{zhang2023remodiffuse}. These participants were instructed to imagine a specific human motion lasting about 10 seconds (equivalent to 120 frames) and subsequently provide an overall textual description (A), detailed textual description (B), and a stickman figure (S) representing a characteristic pose within the imagined motion. The combination of (A, B) was inputted into ReMoDiffuse, while the combination of (A, S) was inputted into StickMotion. Participants then rated the generated results from both methods on a scale of 0 to 10. The final results regarding time consumption and user scores are shown in Tab.~\ref{tab:user}, and the visualizations are presented in Fig.~\ref{fig:user}. Our StickMotion saves users approximately 51.5\% of their time in generating motions consistent with their imagination.

\definecolor{mygray}{gray}{.9}
\begin{table}[t]
\centering
\caption{\small{Comparison between stickman \& text-to-motion and text-to-motion task. ``TA" and ``TB" represent the time cost for overall and detailed descriptions, respectively, while ``TS" denotes the time required for drawing a stickman. Additionally, ``TI'' represents the inference time of the utilized model. }}
\small
\renewcommand\arraystretch{1.2}
\setlength{\tabcolsep}{1.3mm}
\begin{tabular}{lccccccc}
\toprule
 Method & TA & TB & TS & TI & TotalTime$\downarrow$ & Score$\uparrow$    \\ \toprule
\rowcolor{mygray} ReMoDiffuse~\cite{zhang2023remodiffuse} & 8.1   &   24.5   &  -   & 1.2  & 33.8 & 7.3 \\
 StickMotion & 8.1 & -  & 7.7  &  0.7   &  16.4  & 8.5 \\
\bottomrule
\end{tabular}\label{tab:user}
\vspace{-3mm}
\end{table}

\section{Conclusion}
This paper presents a novel stickman condition for motion generation to address users' detailed requirements for body posture through simple textual descriptions. To effectively leverage the stickman condition, we initially developed  an algorithm to generate hand-drawn stickmen based on joint positions from existing datasets. Subsequently, we introduce StickMotion, a diffusion-based approach with multi-condition modules, to create motion sequences that align with textual descriptions and stickmen as conditions while maintaining lower computational complexity. Finally, we propose a new StiSim metric for measuring stickman similarity in traditional text-to-motion evaluation. 
Our objective is to contribute a novel stickman condition to the text-to-motion generation and a Multi-Condition Module for multi-condition scenarios, simplifying and accelerating the process for creators in obtaining desired human motions.


{
    \small
    \bibliographystyle{ieeenat_fullname}
    \bibliography{main}
}

\end{document}